\documentclass[a4paper]{article}

\usepackage{INTERSPEECH2021}
\usepackage{multirow}
\usepackage{balance}
\usepackage{xcolor}

\title{``You don't understand me!":\\ Comparing ASR results for L1 and L2 speakers of Swedish}
\name{Ronald Cumbal$^1$, Birger Moell$^1$, Jos\'{e} Lopes$^2$, Olov Engwall$^1$}
\address{
  $^1$KTH Royal Institute of Technology - Sweden\\
  $^2$Heriot-Watt University - United Kingdom
}
\email{ronaldcg@kth.se, bmoell@kth.se, jd.lopes@hw.ac.uk, engwall@kth.se}

\begin{document}
\maketitle
\begin{abstract}
The performance of Automatic Speech Recognition (ASR) systems has constantly increased in state-of-the-art development. 
However, performance tends to decrease considerably in more challenging conditions (e.g., background noise, multiple speaker social conversations) and with more atypical speakers (e.g., children, non-native speakers or people with speech disorders), which signifies that general improvements do not necessarily transfer to applications that rely on ASR, e.g., educational software for younger students or language learners. 
In this study, we focus on the gap in performance between recognition results for native and non-native, read and spontaneous, Swedish utterances transcribed by different ASR services. We compare the recognition results using Word Error Rate and analyze the linguistic factors that may generate the observed transcription errors.
\end{abstract}


\section{Introduction}
Accompanied by the development of Deep Learning techniques and the increase of computation and data resources, Automatic Speech Recognition (ASR) systems have demonstrated impressive improvement in performance over the last decade \cite{chiu2018state}. 
Many works have already claimed parity with human levels of speech recognition \cite{Xiong2017, Edwards2017}, although this claim may be disputed \cite{Saon2017}. 
However, it is generally understood that these levels of recognition results may only be achieved in specific circumstances. 
For example, high performance levels are often achieved on data obtained from strictly controlled environments (e.g., read audio books \cite{Panayotov_2015}), in well-resourced languages and for speakers that are well represented in the training set.
 
In many real-world applications, these requirements for optimal performance are not met, and the use of ASR may therefore be problematic in applications targeting children \cite{Kennedy2017}, non-native speakers \cite{Ashwell2017,Radzikowski2019} or persons with health conditions affecting their speech (e.g., Parkinson's Disease \cite{Kim2019}).
A setting in which ASR has great potential, but may be vulnerable is educational applications for children and second language (L2) learners, in particular if the offered practice should allow for natural, realistic interaction, e.g., in a school with background noise and multiple speakers.
For educational applications, ASR performance has therefore become a bottleneck and it is customary that studies are instead performed with a human replacing the ASR in a wizard-of-Oz setup \cite{Khalifa2017,engwall2020}.

There are few systematic studies investigating the general understanding that ASR performs poorly for L2 speakers. We in this work explore how three state-of-the-art ASR systems differ in their recognition of first language (L1) and second language speakers.
The comparisons are made between the two sets of speakers, between read sentences and spontaneous social conversational, and between the ASR systems.
The spoken language is Swedish, which may be considered a lower-resourced language when compared to English, further allowing us to qualitatively compare the results with those previously obtained with non-native speakers of English \cite{Ashwell2017,Radzikowski2019}.

The aim of the paper is firstly to determine the extent to which the performance gap in recognition results between L1 and L2 speakers still exists, extend this evaluation on different forms of speech (read vs spontaneous) in a lower resourced language, and formulate how these results may affect the usage of these systems in applications that depend on a specific level of performance to fulfill a task.

\section{Related Work}
Several off-the-shelf or consumer-level implementations of ASR systems were evaluated during the last decade.
In 2013, Morbini \textit{et. al.} \cite{Morbini2013} extended the work of Yao \textit{et. al.} \cite{yao2010} through the evaluation of publicly available recognizers on English dialogue interactions. 
Their findings indicated that different systems reached optimal performance on different dialogue datasets, e.g., Google ASR obtained the best results in the Amani dataset (23.8\% WER) but the worst in Radiobots (36.3\%). As determined by \cite{yao2010}, a considerable variation in performance relates to specific qualities of the datasets (e.g., domain, size and perplexity, out-of-vocabulary rate). 
Kim \textit{et. al.} \cite{Kim2019} compared five online ASR services 
 to assess the quality of their transcriptions in the medical domain. Surprisingly, the Youtube service obtained lower error rates (28\%) in comparison to IBM Watson (50\%). 
Georgila \textit{et. al.} \cite{Georgila2020} investigated the performance of off-the-shelf ASR services on dialogues in different  domains and noisy conditions.
Similarly to the above studies, it was found that the recognizers performed poorly on datasets with specific vocabulary.

While these studies already give an insight into how off-the-shelf ASR technology perform on less optimal datasets, they were all performed with native speakers only. Earlier work on ASR performance with L2 speakers is rare.
Ashwell and Elam \cite{Ashwell2017} evaluated the \textit{Google Web Speech} service with 42 Japanese (and 2 Chinese) learners of English and two native English speakers reading a set of 13 sentences.
When they compared ASR results, the overall accuracy was considerably lower for L2 speakers (65.7\% vs. 89.4\%). 
Furthermore, the most common recognition errors were partly different between L2 and L1 speakers (about 50\% of the errors differed), thus indicating that the misrecognitions were only partly related to the match between the language model and the domain. 
More importantly, it was shown that ASR misrecognition of words uttered by the L2 speakers did not always correspond to pronunciation errors annotated by a human expert, thus indicating that other sources than the acoustic model influence the recognition of L2 speech.
As only 13 read sentences were evaluated, there is a clear need to explore a more extended dataset.
Radzikowski \textit{et. al.} \cite{Radzikowski2019} approached the problem of training ASR systems with limited non-native English speech by employing a technique called Dual Supervised Learning (DSL). 
For this purpose, the authors retrieved YouTube videos of Japanese and Polish speakers pronouncing sentences during English lessons.
The results showed marginal improvements in accuracy, but it serves as a proof of concept for a promising solution to improve training of ASRs with lower data resources. 

In order to better understand the limitations that ASR systems have with respect to non-native speakers, we extend these previous works on quantitative and qualitative data.
We focus on Swedish, as a lower-resourced language, with a dataset comprised of L2 speakers with both spontaneous and read speech, transcribed by three state-of-the-art commercial and research-based ASR systems.
The performance is measured using Word Error Rate (WER) and the linguistic sources of the recognition problems are analyzed.

\begin{table}[b]
  \centering
  \begin{tabular}{cccccc}
    \toprule
    Data set    & Samples           & Speakers  &  Utt. length\\
            & (N/NN)            & (N/NN)     & $\mu$ (sd)\\
    \midrule
    Ville   & 2089 (408/1681)   & 36 (6/30)  & 4.48 (1.63) \\
    CORALL  & 1610 (651/959)    & 30 (6/24)  & 6.90 (6.94) \\
    \bottomrule
\end{tabular}
  \caption{Dataset comparison. N/NN: Native/Non-native ratio. Utt. length: Utterance length measured in number of words.}
  \label{tab:dataset}
\end{table}

\section{Datasets}

In this study we have used two datasets: one with read sentences and a second one with utterances from three-party conversations. Both datasets will be described in this section with a detailed comparison between them presented in Table \ref{tab:dataset}.

\subsection{Read sentences (Ville)}
The dataset with read sentences was produced as part of a virtual teacher program for L2 learners of Swedish ''Ville'' \cite{wik2009}. 
The study included participants who self-reported an \textit{Advanced Beginner} level of speaking proficiency. 
These participants had 18 different native languages, of which French, German and Chinese were the most frequent. 
The system employed both perception and production exercises to train the learners. 
In the latter, participants were guided and evaluated on reading sentences shown on flashcards (e.g., ``förstå'' [\textit{understand}], ``och toalettpapper'' [\textit{and toilet paper}] ``kaffe och te'' [\textit{coffe and tea}], ``och köpa lite mat'' [\textit{and buy some food}]). 
The dataset is formed with these card-texts and speech recordings pairs. 
Since during the experiment some of the participants interacted in various practice sessions, we filtered the dataset to include only one sample of each read text from each participant.

\subsection{Social conversations (CORALL)}
The second dataset we use in this study was collected during experiments of social conversation practice with L2 learners of Swedish led by a robot peer \cite{engwall2020} or a native speaker. 
In the former, pairs of participants interacted with a robot (in a wizard-of-Oz set-up) that employed different conversation style strategies and the two participants in each session were recorded with individual headset microphones. 
In the latter, as part of pilot studies of the same experiment, a human speaker directed the conversation in the same social conversation setting.
The language learners self-reported a level of \textit{Basic} to \textit{Intermediate} level of Swedish proficiency. 
These recordings were transcribed by Swedish native speakers, with each annotator focusing on separate sections, and the audio files were segmented per speaker. Common disfluencies in Swedish language, e.g., "ehm", "ahm", "hm", were disregarded, and word fragments, e.g., "fö-", "framt-", were transcribed as complete words.

\section{Speech Recognizers}\label{sec:Recognizers}
As a lower-resource language, Swedish is not included in many prevalent ASR or Speech-To-Text (STT) systems (e.g. Amazon Transcribe\footnote{https://aws.amazon.com/transcribe/}, IBM Watson\footnote{https://www.ibm.com/se-en/cloud/watson-text-to-speech},  Houndify\footnote{https://www.houndify.com/static-faq}, VOSK\footnote{https://alphacephei.com/vosk/}). 
In our search, we found three off-the-shelf ASR services that could process Swedish speech, but only two had an API available at the moment of developing this work (Trint\footnote{https://trint.com/} now offers an API for dictations). 
Furthermore, we used an open source Swedish ASR model available through the platform Huggingface\footnote{https://huggingface.co/}. The final selection includes the following ASRs:

\subsection{Google Cloud}
Google's speech recognition is provided through the \textit{Speech-to-Text API}\footnote{https://cloud.google.com/speech-to-text}. 
Improvements on the recognition process can be performed by defining a set of expected words (\textit{Speech Adaptation}) and applying specific models that match the audio characteristics (\textit{Domain-specific models}). 

\subsection{Microsoft Azure}
Microsoft presents the \textit{Speech-to-text} service as part of the Azure project. 
Their ASR technology is build for conversational and dictation scenarios with the possibility to train custom acoustic, language, and pronunciation models (e.g. language model through the \textit{Custom Speech} tool). 


\subsection{Huggingface}
We used the Huggingface transformers library \cite{wolf-etal-2020-transformers} to test the recent released of the Wav2vec2 model architecture by Baevski \textit{et al.} \cite{baevski2020wav2vec}. The Swedish  model\footnote{KBLab/wav2vec2-large-xlsr-53-swedish} was trained by researchers at the Swedish Royal Library, who had previously worked on a Swedish version of the BERT language model. The Wav2Vec2 model learns audio embeddings through training on masking parts of the audio, similar to methods used in text models that masks words in sentences during the training stage \cite{devlin2018bert}. This approach allows to use pre-trained models and fine-tune them with lesser amounts of data to still achieve competitive results.

\section{Evaluation}
\subsection{Performance}
The datasets were transcribed with the default options suggested in each API of the Google and Microsoft Azure ASRs. 
No additional processing steps, e.g., tuning acoustic models or extending language models, were taken. 
In the case of the Huggingface ASR, the model was trained on the Swedish NST dataset \cite{NST_dataset} producing a WER of 16.94\% when evaluated on the Common Voice dataset \cite{commonvoice2020}.
We evaluate the performance of each ASR using the Word Error Rate metric and removed punctuation marks, converted numerical tokens to text, and transformed all transcriptions to lower-case in this process. 
Furthermore, as the processing of some audios did not produce any transcription at all, we counted these occurrences as the Number of samples that Failed to be Recognized (NFR). 
The results for WER and NFR are also evaluated by different levels of number of words contained in each sample. We segment these levels in short utterances (S), containing samples with 4 words or less, medium utterances (M) that range from 5 to 10 words, and long utterances (L) that contain utterances with more than 10 words.
We perform this division since it is usually understood that shorter utterances tend to have worse recognition results (for native speakers) and we would like to analyze the impact of utterance length when considering non-native speech.

\begin{table}[t]
  \centering
  \begin{tabular}{ccccc}
    \toprule
    Dataset     &    Speech   &  Goo. & Mic. & Hug. \\
    \midrule
                          & Native     & 0.162 & 0.111 & 0.522 \\
                            & Non-native & 0.325 & 0.410 & 0.593 \\
                            \cline{2-5}
    Ville                  & Spanish    & 0.483 & 0.597 & 0.744 \\
    (Read sentences)        & Chinese    & 0.431 & 0.552 & 0.687 \\
                            & French     & 0.378 & 0.487 & 0.733 \\
                            &  \vdots  & \vdots & \vdots & \vdots \\
                            & Italian & 0.201 & 0.2347 & 0.321 \\
    \midrule
                            & Native  & \textbf{0.412} & 0.356 & \textbf{0.641} \\
                            & Non-native & \textbf{0.421} & 0.507 & \textbf{0.663} \\
                            \cline{2-5}
                            & French & 0.528 & 0.784 & 0.743 \\
    CORALL                  & Spanish  & 0.451 & 0.581 & 0.726 \\
    (Social conv.)          & Chinese & 0.504 & 0.525 & 0.693 \\
                            &  \vdots & \vdots&\vdots & \vdots\\
                            & Punjabi & 0.280 & 0.320 & 0.556 \\
                            & Italian & 0.335 & 0.310 & 0.276 \\
    \bottomrule
  \end{tabular}
  \caption{Word Error Rate (WER) across platforms, dataset, and language of speaker. Bold values indicate unexpected results between non-native and native recognition results. WERs are presented for particular L2 results in descending order.}
  \label{tab:wer_results}
\end{table}

\subsection{Transcription Errors}\label{sec:transcription_errors}
The evaluation of the produced transcriptions was focused on the most common misrecognized words, i.e., Substituted and Deleted words. 
We employ two metrics in this step. 
First, we count the number of error words and divide this value by the number of times it appears in the dataset, defined as $ef_{w}$. 
Since this metric may accentuate words that appear very few times and are misrecognized in these few instances, we compute a normalized version by subtracting the frequency of the word from the number of words in the complete dataset. We refer to this metric as normalized frequency $enf_{w}$ in Equation \ref{eq:error_importance}. 

\begin{table}[t]
  \centering
  \begin{tabular}{cccccc}
    \toprule
     & \multirow{2}{*}{Speech} & \multirow{2}{*}{Size} & \multicolumn{3}{c}{WER (NFR)} \\
                            \cline{4-6}
                &            &  & Goo. & Mic. & Hug.  \\
    \midrule
    \multirow{6}{*}{V} & \multirow{2}{*}{N}    & S & 0.24 (5) & 0.15 (0) & 0.78 (0)\\
                            &                    & M & 0.11 (2) & 0.09 (2) & 0.33 (1)\\
                            \cline{2-6}
                            & \multirow{2}{*}{NN} & S & 0.32 (3) & 0.54 (1) & 0.82 (1)\\
                            &                    & M & \textbf{0.33 (4)} & 0.37 \textbf{(7)} & 0.52 (4)\\
    \midrule
    \multirow{6}{*}{C} &                      & S & 0.50 (81) & 0.37 (63) & 0.82 (10)\\
                            & N                 & M & 0.38 (8) & 0.31 (8) & 0.51 (1)\\
                            &                   & L & 0.35 (0) & \textbf{0.42} (0) & 0.46 (1) \\
                            \cline{2-6}
                            &                   & S & 0.48 (93) & 0.51 (47) & 0.79 (14)\\
                            & NN             & M & 0.38 (4) & 0.46 (2) & 0.62 (1) \\
                            &                   & L & \textbf{0.46} (0)  & \textbf{0.61} (1) & 0.57 (0) \\
    \bottomrule
  \end{tabular}
  \caption{Analysis at utterance length level. WER: Word Error Rate and NFR: Number of full samples Failed to Recognize. 
  \\ V: Ville; C: CORRAL; N: Native; NN: Non-Native. S: Short; M: Medium; L: Long. Bold values show unexpected high results.}
  \label{tab:word_error}
\end{table}

\begin{table*}[t]
  \centering
  \begin{tabular}{c|l|l|l|l|l|l}
    \toprule
    & \multicolumn{3}{c|}{ASR Errors Normalized Frequency $enf_{w}$} &  \multicolumn{3}{c}{ASR Errors Frequency $ef_{w}$} \\
    \midrule
	& Non-native  & Both	& Native	& Non-native	& Both &	Native \\
    \midrule
    Deletions   & ah, förstår, min,   & också,  hur, att, så, & då, ett	& hip, låna	&   &där, ganska, gann \\
                & mycket, nej, om,    & i, man, för, är, här,   &   &  & \\
                & vill, väder         &  ja, jag, sverige, och, &   &  &   & \\
                &                     & vad, du, med, det, på   &   &   &   &\\
    \midrule
    Substitutions &	bor, därför, familj,   & att, till,   & lär,  & jättesnabbt,  & & er, gjorde \\
                    &  för, förstår, min, & är, vad, du, & lära, ni,     & komedier, serie,    & & ihåg, sand, store,\\
                    &  mycket, repetera   & det, en      &  språkcafé     & skillnad         & & såna, trevligt, ute\\
    \bottomrule
  \end{tabular}
  \caption{Words that all ASRs had problems with based on whether the errors pertain to only non-native speakers, only native speakers, or shared by both. }
  \label{tab:transcription_results}
\end{table*}


\begin{equation}
  enf_{w} = \frac{{num\_errors}_{w}}{{num\_words}_{dataset} - num\_occurrence_{w}}
  \label{eq:error_importance}
\end{equation}

\section{Results}
\subsection{Performance}
As expected, and presented in Table~\ref{tab:wer_results}, the samples corresponding to native speakers tend to have a lower WER, but this difference is less obvious when  utterances correspond to spontaneous speech. 
In this dataset, the only statistically significant result is observed in the transcriptions generated by the Microsoft ASR (N: $0.36$ vs. NN: $0.51$,  $p<0.05$), while the other ASRs fail to show a significant difference (Google N: $0.41$ vs. NN: $0.42$ and Huggingface N: $0.64$ vs. NN: $0.66$).
All computation of statistical significance is done through a Welch’s t-test of the average WER per speaker.
When we analyze the mother tongue of the non-native speakers, we find that most L1 speakers (e.g., Spanish, Chinese, and French) tend to have a worse WER when compared to Swedish speakers. 
However, it's interesting to see that, in both datasets, samples from Italian speakers have results comparable, or better, than the native Swedish speaker. Since our dataset includes few Italian speakers, we cannot assess the significance of this finding. 

The performance results at different utterance lengths are shown in  Table \ref{tab:word_error}.  
If we analyze the read sentences dataset, we find that the recognition results are indeed better for the native speakers' medium-long utterances than the short (about half of the WER), but that the differences are smaller for the non-native speakers, and that the Google ASR presents almost no difference when comparing short and medium utterances (M: $0.33$ vs. S: $0.32$). 
Further, when we evaluate the spontaneous speech, the previous pattern is also perceived with the native speech, except for the results of the Microsoft ASR (L: $0.42$ $>$ S: $0.37$, $p<0.05$). At first glance, it seems that the results in the spontaneous non-native speech also have an opposite behavior than expected, as the longer utterances have similar or worse WER when comparing with the short ones. 
However, when considering the number of non-recognized samples for short utterances, NFR, this observation changes. 
These results show that for all spontaneous speech, the ASRs frequently fail to produce a transcription for short utterances (Google: 81+93, Microsoft: 63+47 and Huggingface: 10+14). 
Nonetheless, we do note that for two ASRs, the longer utterances still have a higher WER when compared to medium-length ones (Google L: $0.46$ $>$ M: $0.38$, $p$=0.39; and Microsoft L: $0.61$ $>$ M: $0.46$, $p<0.05$), which is not the expected outcome.

\subsection{Transcription Errors}
To evaluate the type of word errors that commonly occur in misrecognized utterances, we used the metrics from Section \ref{sec:transcription_errors} to score deleted or substituted words. This analysis is performed over the transcriptions of all ASRs to determine patterns shared by all of these systems. We focus on the spoken conversations dataset to place more emphasis on applications where the ASR is used for highly interactive tasks.  
Table \ref{tab:transcription_results} shows the top most common errors when ranked by frequency ($ef_{w}$) and normalized frequency of occurrence ($enf_{w}$). 
We grouped these error words on whether they are only seen in samples from non-native speakers, only native speakers or present in both.  
The overview of these errors further supports the assumption that very short words (monosyllables) will often be misrecognized by ASR systems. This is most notable in the words misrecognized from both non-native and native speakers, e.g. ``ja'' [\textit{yes}], ``och'' [\textit{and}], ``du'' [\textit{you}], ``jag'' [\textit{I}].
It is also important to notice that common errors for only non-native speakers include words like 
``förstår'' [\textit{understand}] and ``repetera'' [\textit{repeat}], which is problematic, since these are used by learners to signal non-understanding or request a repetition. 
The word ``jättesnabbt" [\textit{very fast}] in the right part of the table also belongs to this category, since it was used by learners to signal that the robot was speaking too fast.
For the native speakers, the most notable problematic word is ``språkcaf\'e" [\textit{Language Caf\'e}], which is the term used to describe a specific setting for practice conversation for language learners (that was replicated with the robot set-up).
Finally, the words displayed in the right side of Table \ref{tab:transcription_results}, corresponding to the most common errors divided by their frequency in the datasets (e.g., ``komedier" [\textit{comedies}]) are often closely related to the topics of social conversations.

\section{Discussion}

The results of comparing the transcriptions generated by common ASRs services on native and non-native speakers confirm the assumption that Word Error Rates increase (to almost the double rate) with L2 speakers for read sentences
However, with the spontaneous speech dataset, we found that the performance of two ASRs (Google and Huggingface) deteriorates to similar levels for both native and non-natives speakers.
The fact that Microsoft Azure ASR performs better for native speakers may be related to the development of the system, as it is built for conversations. These results strengthen the notion that it is important that common ASR services are applied in conditions for which they have been trained. 

To expand our analysis, we evaluated the recognition performance at utterance-length, where we expected to find lower Word Error Rates for longer samples of speech, as the language model should provide additional information to improve the recognition. 
This assumption was not fully supported, as we found that -- for the non-native speakers -- medium (in read sentences) and longer utterances (in spontaneous speech) performed similar to or worse than shorter and medium utterances, respectively. 
However, we found that fully failed recognition outputs (i.e. NFR) were more frequent for the shorter utterances in all cases.
These findings are specifically concerning if the ASR systems are employed for less controlled interactions with non-native speaker, where, on the one hand, simple expressions like "yes" or "what?" may not generate any recognition at all, and on the other, longer expression, like opinions or ideas, may contain too many errors that cripple any further interpretation.  

In particular, we find that educational applications oriented to language learning, where computer or robot assistants expand the possible practice activities, may be limited in which interaction they may have with students. 
For example, when we analyzed the word errors of the most frequent misrecognized utterances, we found that for non-native speakers, there were specific misrecognized words ("understand" and "repeat") that signal important states of uncertainty in the user.

We do acknowledge that our study has some notable limitations. 
None of our systems were adapted (fine tuned) for non-native speakers, as this step requires much more data.
This process is part of our future work, along with investigating the use of large language models and collect additional data. 
Finally, related to the notable results of the Italian participants, although unlikely, it's possible that the Italian participants had a linguistic level very close to native speakers', or that both languages have a similar phoneme set, but further research is required.


\balance

\section{Conclusion}
This study expands the comparison of L1 and L2 speech when processed by ASRs. We found more support for poorer results generated from non-native speech, accentuated for spontaneous speech, and indicate how particular recognition errors, like essential words or long utterances, reduce the usability of these systems in important (education), but less controlled, scenarios. 

\section{Acknowledgements}
This work was supported by the Swedish Research Council Grant 2016-03698 ``Collaborative Robot Assisted Language Learning" and the VINNOVA grant 2019-02997 ``AI teacher for more efficient language learning''.

\bibliographystyle{IEEEtran}

\bibliography{ASR,mybib}

\begin{thebibliography}{10}
\providecommand{\url}[1]{#1}
\csname url@samestyle\endcsname
\providecommand{\newblock}{\relax}
\providecommand{\bibinfo}[2]{#2}
\providecommand{\BIBentrySTDinterwordspacing}{\spaceskip=0pt\relax}
\providecommand{\BIBentryALTinterwordstretchfactor}{4}
\providecommand{\BIBentryALTinterwordspacing}{\spaceskip=\fontdimen2\font plus
\BIBentryALTinterwordstretchfactor\fontdimen3\font minus
  \fontdimen4\font\relax}
\providecommand{\BIBforeignlanguage}[2]{{%
\expandafter\ifx\csname l@#1\endcsname\relax
\typeout{** WARNING: IEEEtran.bst: No hyphenation pattern has been}%
\typeout{** loaded for the language `#1'. Using the pattern for}%
\typeout{** the default language instead.}%
\else
\language=\csname l@#1\endcsname
\fi
#2}}
\providecommand{\BIBdecl}{\relax}
\BIBdecl

\bibitem{chiu2018state}
C.-C. Chiu, T.~N. Sainath, Y.~Wu, R.~Prabhavalkar, P.~Nguyen, Z.~Chen,
  A.~Kannan, R.~J. Weiss, K.~Rao, E.~Gonina \emph{et~al.}, ``State-of-the-art
  speech recognition with sequence-to-sequence models,'' in \emph{2018 IEEE
  International Conference on Acoustics, Speech and Signal Processing
  (ICASSP)}.\hskip 1em plus 0.5em minus 0.4em\relax IEEE, 2018, pp. 4774--4778.

\bibitem{Xiong2017}
W.~Xiong, J.~Droppo, X.~Huang, F.~Seide, M.~L. Seltzer, A.~Stolcke, D.~Yu, and
  G.~Zweig, ``{Toward Human Parity in Conversational Speech Recognition},''
  \emph{IEEE/ACM Trans. Audio Speech Lang. Process.}, vol.~25, no.~12, pp.
  2410--2423, 2017.

\bibitem{Edwards2017}
E.~Edwards, W.~Salloum, G.~P. Finley, J.~Fone, G.~Cardiff, M.~Miller, and
  D.~Suendermann-Oeft, ``Medical speech recognition: Reaching parity with
  humans,'' in \emph{Speech and Computer}, A.~Karpov, R.~Potapova, and
  I.~Mporas, Eds.\hskip 1em plus 0.5em minus 0.4em\relax Cham: Springer
  International Publishing, 2017, pp. 512--524.

\bibitem{Saon2017}
G.~Saon, G.~Kurata, T.~Sercu, K.~Audhkhasi, S.~Thomas, D.~Dimitriadis, X.~Cui,
  B.~Ramabhadran, M.~Picheny, L.~L. Lim, B.~Roomi, and P.~Hall, ``{English
  conversational telephone speech recognition by humans and machines},''
  \emph{Proc. Annu. Conf. Int. Speech Commun. Assoc. INTERSPEECH}, vol.
  2017-August, pp. 132--136, 2017.

\bibitem{Panayotov_2015}
V.~{Panayotov}, G.~{Chen}, D.~{Povey}, and S.~{Khudanpur}, ``Librispeech: An
  asr corpus based on public domain audio books,'' in \emph{2015 IEEE
  International Conference on Acoustics, Speech and Signal Processing
  (ICASSP)}, 2015, pp. 5206--5210.

\bibitem{Kennedy2017}
\BIBentryALTinterwordspacing
J.~Kennedy, S.~Lemaignan, C.~Montassier, P.~Lavalade, B.~Irfan,
  F.~Papadopoulos, E.~Senft, and T.~Belpaeme, ``Child speech recognition in
  human-robot interaction: Evaluations and recommendations,'' in
  \emph{Proceedings of the 2017 ACM/IEEE International Conference on
  Human-Robot Interaction}, ser. HRI '17.\hskip 1em plus 0.5em minus
  0.4em\relax New York, NY, USA: Association for Computing Machinery, 2017, p.
  82–90. [Online]. Available: \url{https://doi.org/10.1145/2909824.3020229}
\BIBentrySTDinterwordspacing

\bibitem{Ashwell2017}
T.~Ashwell and J.~R. Elam, ``How accurately can the \expandafter{Google} web
  speech \expandafter{API} recognize and transcribe \expandafter{Japanese}
  \expandafter{L2} \expandafter{English} learners’ oral production?''
  vol.~13, 2017, pp. 59--76.

\bibitem{Radzikowski2019}
\BIBentryALTinterwordspacing
K.~Radzikowski, R.~Nowak, L.~Wang, and O.~Yoshie, ``Dual supervised learning
  for non-native speech recognition,'' \emph{EURASIP Journal on Audio, Speech,
  and Music Processing}, 2019. [Online]. Available:
  \url{https://doi.org/10.1186/s13636-018-0146-4}
\BIBentrySTDinterwordspacing

\bibitem{Kim2019}
\BIBentryALTinterwordspacing
J.~Y. Kim, C.~Liu, R.~A. Calvo, K.~McCabe, S.~C.~R. Taylor, B.~W. Schuller, and
  K.~Wu, ``{A Comparison of Online Automatic Speech Recognition Systems and the
  Nonverbal Responses to Unintelligible Speech},'' pp. 1--13, 2019. [Online].
  Available: \url{http://arxiv.org/abs/1904.12403}
\BIBentrySTDinterwordspacing

\bibitem{Khalifa2017}
A.~Khalifa, T.~Kato, and S.~Yamamoto, ``Measuring effect of repetitive queries
  and implicit learning with joining-in type robot assisted language learning
  system,'' in \emph{ISCA workshop on Speech and Language Technology in
  Education}, 2017, pp. 13--17.

\bibitem{engwall2020}
O.~Engwall, J.~Lopes, and A.~Åhlund, ``Robot interaction styles for
  conversation practice in second language learning,'' \emph{International
  Journal of Social Robotics}, 2020.

\bibitem{Morbini2013}
F.~Morbini, K.~Audhkhasi, K.~Sagae, R.~Artstein, D.~Can, P.~Georgiou,
  S.~Narayanan, A.~Leuski, and D.~Traum, ``{Which \expandafter{ASR} should i
  choose for my dialogue system?}'' \emph{SIGDIAL 2013 - 14th Annu. Meet. Spec.
  Interes. Gr. Discourse Dialogue, Proc. Conf.}, no. August, pp. 394--403,
  2013.

\bibitem{yao2010}
X.~Yao, P.~Bhutada, K.~Georgila, K.~Sagae, R.~Artstein, and D.~Traum,
  ``{Practical evaluation of speech recognizers for virtual human dialogue
  systems},'' \emph{Proc. 7th Int. Conf. Lang. Resour. Eval. Lr. 2010}, pp.
  1597--1602, 2010.

\bibitem{Georgila2020}
K.~Georgila, A.~Leuski, V.~Yanov, and D.~Traum, ``{Evaluation of Off-the-shelf
  Speech Recognizers Across Diverse Dialogue Domains},'' no. May, pp.
  6469--6476, 2020.

\bibitem{wik2009}
P.~Wik, R.~Hincks, and J.~Hirschberg, ``Responses to \expandafter{Ville}: a
  virtual language teacher for swedish,'' in \emph{SLaTE}, 2009.

\bibitem{wolf-etal-2020-transformers}
\BIBentryALTinterwordspacing
T.~Wolf, L.~Debut, V.~Sanh, J.~Chaumond, C.~Delangue, A.~Moi, P.~Cistac,
  T.~Rault, R.~Louf, M.~Funtowicz, J.~Davison, S.~Shleifer, P.~von Platen,
  C.~Ma, Y.~Jernite, J.~Plu, C.~Xu, T.~L. Scao, S.~Gugger, M.~Drame, Q.~Lhoest,
  and A.~M. Rush, ``Transformers: State-of-the-art natural language
  processing,'' in \emph{Proceedings of the 2020 Conference on Empirical
  Methods in Natural Language Processing: System Demonstrations}.\hskip 1em
  plus 0.5em minus 0.4em\relax Online: Association for Computational
  Linguistics, Oct. 2020, pp. 38--45. [Online]. Available:
  \url{https://www.aclweb.org/anthology/2020.emnlp-demos.6}
\BIBentrySTDinterwordspacing

\bibitem{baevski2020wav2vec}
A.~Baevski, H.~Zhou, A.~Mohamed, and M.~Auli, ``wav2vec 2.0: A framework for
  self-supervised learning of speech representations,'' \emph{arXiv preprint
  arXiv:2006.11477}, 2020.

\bibitem{devlin2018bert}
J.~Devlin, M.-W. Chang, K.~Lee, and K.~Toutanova, ``Bert: Pre-training of deep
  bidirectional transformers for language understanding,'' \emph{arXiv preprint
  arXiv:1810.04805}, 2018.

\bibitem{NST_dataset}
``Nst swedish asr database,'' \url{https://www.nb.no/sprakbanken/en}, accessed:
  2021-06-14.

\bibitem{commonvoice2020}
R.~Ardila, M.~Branson, K.~Davis, M.~Henretty, M.~Kohler, J.~Meyer, R.~Morais,
  L.~Saunders, F.~M. Tyers, and G.~Weber, ``Common voice: A
  massively-multilingual speech corpus,'' in \emph{Proceedings of the 12th
  Conference on Language Resources and Evaluation (LREC 2020)}, 2020, pp.
  4211--4215.

\end{thebibliography}

\end{document}